\documentclass[sigconf]{acmart}
\usepackage{multirow}
\usepackage{graphicx}
\usepackage[normalem]{ulem}
\useunder{\uline}{\ul}{}
\usepackage{enumitem}

\AtBeginDocument{%
  }

\copyrightyear{2026}
\acmYear{2026}
\setcopyright{cc}
\setcctype{by}
\acmConference[WWW '26] {Proceedings of the ACM Web Conference 2026}{April 13--17, 2026}{Dubai, United Arab Emirates.}
\acmBooktitle{Proceedings of the ACM Web Conference 2026 (WWW '26), April 13--17, 2026, Dubai, United Arab Emirates}
\acmISBN{979-8-4007-2307-0/2026/04}
\acmDOI{10.1145/XXXXXX.XXXXXX}

\begin{document}

\title{Towards Token-Level Text Anomaly Detection}

\author{Yang Cao}
\affiliation{%
  \institution{Great Bay University}
  \institution{Tsinghua University}
  \city{Dongguan}
  \state{Guangdong}
  \country{China}
}
\email{charles.cao@ieee.org}


\author{Bicheng Yu}
\affiliation{%
  \institution{Great Bay University}
  \institution{Shenzhen University}
  \city{Dongguan}
  \state{Guangdong}
  \country{China}
}
\email{242433004@stu.gbu.edu.cn}

\author{Sikun Yang}
\authornote{Corresponding author.}
\affiliation{%
  \institution{Great Bay University}
 \institution{Dongguan Key Laboratory for AI and Dynamical Systems}
  \city{Dongguan}
  \state{Guangdong}
  \country{China}
}
\email{sikunyang@gbu.edu.cn}

\author{Ming Liu}
\affiliation{%
  \institution{Deakin University}
  \city{Melbourne}
  \state{VIC}
  \country{Australia}
}
\email{m.liu@deakin.edu.au}

\author{Yujiu Yang}
\affiliation{%
  \institution{Tsinghua University}
  \city{Shenzhen}
  \state{Guangdong}
  \country{China}
}
\email{yang.yujiu@sz.tsinghua.edu.cn}




\begin{abstract}
  Despite significant progress in text anomaly detection for web applications such as spam filtering and fake news detection, existing methods are fundamentally limited to document-level analysis, unable to identify which specific parts of a text are anomalous. We introduce token-level anomaly detection, a novel paradigm that enables fine-grained localization of anomalies within text. We formally define text anomalies at both document and token-levels, and propose a unified detection framework that operates across multiple levels. To facilitate research in this direction, we collect and annotate three benchmark datasets spanning spam, reviews and grammar errors with token-level labels. Experimental results demonstrate that our framework get better performance than other 6 baselines, opening new possibilities for precise anomaly localization in text. All the codes and data are publicly available on \url{https://github.com/charles-cao/TokenCore}.
\end{abstract}



\maketitle


\section{Introduction}

Text anomaly detection has become increasingly important across diverse web applications including spam filtering, fake news detection, and machine-generated content recognition. Despite significant progress in these areas, existing methods operate at two extremes: either document-level detection that classifies entire texts without locating anomalous regions, or highly specialized token-level tasks such as spelling correction, grammatical error detection, or named entity recognition (NER)~\cite{li2020survey} that target only specific, predefined anomaly types. This high-level anomaly detection leaves a significant gap in the field. Document-level methods sacrifice interpretability, users cannot understand why a text is flagged as anomalous or locate the problematic elements. Meanwhile, specialized token-level tasks require task-specific training data and cannot generalize across different anomaly types. A spelling checker cannot detect semantic inconsistencies, and a grammar corrector cannot identify character corruption. In practice, however, texts may contain diverse anomalies simultaneously, and a unified framework being capable of detecting and localizing arbitrary anomaly types remains absent.


To address this gap, we propose that text anomaly detection should advance to token-level granularity with a unified framework capable of detecting and localizing anomalous tokens regardless of the specific anomaly type—be it orthographic errors, semantic deviations, or syntactic violations~\cite{cao-etal-2025-text}. Achieving this unified token-level detection presents unique challenges. First, the extreme class imbalance at token-level where anomalous tokens may constitute less than 1\% of all tokens even in anomalous documents—makes detection significantly harder than document-level detection. Second, pre-trained language models are optimized for semantic similarity rather than anomaly sensitivity, potentially smoothing out surface-level anomalies such as spelling errors. Third, aggregating from token-level to document-level scores requires preserving anomaly signals without amplifying false positives.

The contributions of this paper are: \textbf{1)} first introducing token-level text anomaly detection that localizes anomalous tokens regardless of anomaly types, bridging the gap between coarse-grained document-level detection and narrow task-specific approaches. \textbf{2)} labeling and releasing three real-world datasets covering character corruption, semantic anomalies, and grammatical errors with token-level annotations, enabling systematic evaluation of fine-grained detection. \textbf{3)} proposing TokenCore which constructs a memory bank of normal token embeddings and uses nearest neighbor matching for anomaly scoring. \textbf{4)} demonstrating that TokenCore achieves the best average performance across different anomaly types at both token and document-levels. \textbf{5)} summarizing the existing key challenges for token-level text anomaly detection. 

\section{Related Work}

\textbf{Text Embeddings.} Text representation has evolved from sparse to dense word embeddings~\cite{pennington2014glove}. While these static embeddings capture semantic relationships through co-occurrence patterns, they fail to account for contextual variations in word usage. Transformer-based models addressed this limitation by producing contextualized representations through bidirectional attention mechanisms, enabling words to have different embeddings based on surrounding context. These pre-trained language models have become the standard backbone for various NLP tasks, providing rich semantic representations that can be adapted for downstream applications.

\textbf{Anomaly Detection Methods.} Traditional anomaly detection algorithms can be broadly categorized into density-based, isolation-based, statistical, and deep learning approaches. Density-based methods like LOF~\cite{breunig2000lof} identify anomalies by measuring local density deviations relative to neighboring points. Isolation-based methods~\cite{cao2025anomaly} such as Isolation Forest~\cite{liu2008isolation} exploit the property that anomalies are sparse and easier to isolate through random partitioning. Statistical approaches including ECOD~\cite{li2022ecod} detect anomalies through their deviation from estimated data distributions using cumulative distribution functions. Deep learning methods like Deep SVDD~\cite{ruff2018deep} and AutoEncoder learn compact representations of normal data and identify anomalies as points with high reconstruction error or large distance from the learned hypersphere. More recent methods like LUNAR~\cite{goodge2022lunar} combine nearest neighbor information with neural networks to capture complex anomaly patterns. These methods operate on feature representations and are agnostic to the data domain, making them applicable to text anomaly detection when combined with appropriate text embeddings.

\vspace{-14pt}

\section{What are anomalies in text?}

A fundamental question remains: \textbf{what specific phenomena in text can be considered anomalous?} 
We categorize text anomalies into document-level and token-level phenomena.

\textbf{Document-level anomalies} encompass abnormalities that manifest at the broader textual structure and meaning level. \textbf{1) Syntactic anomalies} involve violations of grammatical rules and sentence structure, such as missing sentence components, disordered word sequences, or systematic grammar rule violations that impede comprehension. \textbf{2) Semantic anomalies} occur when the meaning or logical coherence of the text is compromised, including logical contradictions within the content, inconsistent themes or topics, and confused semantic relationships between concepts or sentences. \textbf{3) Pragmatic anomalies} relate to the appropriateness of language use within specific contexts, such as inappropriate register or style for the intended audience, sudden stylistic shifts within a document, or language use that violates contextual expectations. \textbf{4) Structural anomalies} involve problems with the organization and format of the document, including disordered paragraph organization, missing logical flow in argumentation, and non-standard formatting that deviates from expected conventions.

\textbf{Token-level anomalies} focus on abnormalities at the lexical and sub-lexical levels. \textbf{1) Lexical anomalies} involve inappropriate word choices or usage, such as the use of extremely rare or archaic vocabulary inappropriate for the context, misuse of technical terminology, and inappropriate word collocations that violate conventional usage patterns. \textbf{2) Orthographic anomalies} encompass spelling and formatting irregularities, including spelling errors, inconsistent capitalization, and misuse of special characters or punctuation. \textbf{3) Morphological anomalies} involve incorrect word formation or inflection, such as tense errors, singular-plural disagreements, and incorrect word derivations.

We formalize these anomalies from a distributional perspective. Let $\mathcal{D}_n = \{d_1, d_2, \ldots, d_N\}$ denote a corpus of $N$ normal documents, and let $P_n$ represent the underlying distribution of normal text.

\textbf{Definition 1 (Token-Level Anomaly).} Given a mapping function $\phi: \mathcal{W} \rightarrow \mathbb{R}^d$ that embeds tokens into a $d$-dimensional space, a token $w$ is anomalous if its embedding $\phi(w)$ deviates significantly from the distribution of normal token embeddings. 

\textbf{Definition 2 (Document-Level Anomaly).} A document $d = (w_1, w_2, \ldots, w_T)$ is anomalous if: (1) it contains at least one anomalous token, or (2) its document-level representation deviates from the normal document distribution. 

\vspace{-10pt}
\section{Datasets}

To facilitate research in token-level text anomaly detection, we construct three benchmark datasets spanning different types of textual anomalies: corrupted characters, semantic anomalies in reviews, and grammatical errors. Each dataset is carefully curated to include both document-level and token-level annotations, enabling fine-grained anomaly localization. 

\textbf{SMS Spam Dataset}\footnote{\scriptsize{\url{https://huggingface.co/datasets/kendx/NLP-ADBench/tree/main/datasets/sms_spam}}} focuses on character-level anomalies manifested as corrupted or gibberish text sequences. Specifically, we randomly select portions of normal messages and inject meaningless character sequences such as `jasdhnjsdb' to simulate real-world scenarios where text corruption occurs due to encoding errors, transmission issues, or deliberate obfuscation attempts in spam messages. This dataset has 393 anomalous samples in total 4518 documents, anomaly rate is 8.7\%. 

\textbf{Restaurant Review Dataset} is collected from Google Maps reviews a restaurant located in USA. This dataset targets semantic anomalies within customer reviews, specifically focusing on negative sentiment expressions embedded in review text. We manually inspect each review and apply the following annotation criteria: positive reviews without grammatical errors are labeled as normal, while negative reviews containing explicit negative sentiment indicators are labeled as anomalous. For token-level annotation, we identify and mark specific words or phrases that convey strong negative sentiment, such as criticism, complaints, or disapproval. These marked tokens represent semantic anomalies within the text community of restaurant reviews, where predominantly positive or neutral expressions constitute the norm. This dataset has 50 anomalous samples in total 1100 documents, anomaly rate is 4.5\%. 

\textbf{Grammar Correction Dataset}\footnote{\scriptsize{\url{https://www.kaggle.com/datasets/satishgunjal/grammar-correction?resource=download}}} from Kaggle provides paired examples of grammatically correct and incorrect sentences. We construct our evaluation set by selecting 270 correct sentences as normal samples and 30 incorrect sentences as anomalous samples, maintaining an imbalanced distribution typical of real-world anomaly detection scenarios.
This dataset has 30 anomalous samples in total 300 documents, anomaly rate is 10\%.

\section{Memory Bank Framework}

\begin{figure}[!htbp]
    \centering
    \includegraphics[width=0.7\linewidth]{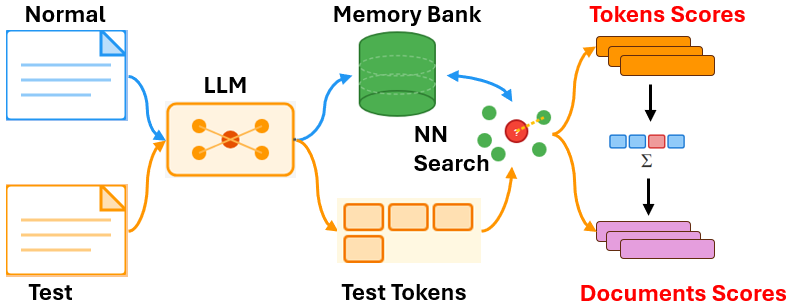}
    \caption{Demonstration of the key steps for TokenCore. Normal documents are encoded by LLM to construct a memory bank of normal token embeddings. During testing, test documents are tokenized and each token embedding is compared against the memory bank via nearest neighbor search. Token-level anomaly scores are computed based on distances to the nearest normal neighbors, and document-level scores are obtained through mean aggregation.}
    \label{fig:framework}
\end{figure}

Figure~\ref{fig:framework} illustrates the overall architecture. TokenCore adapts the memory bank paradigm from PatchCore~\cite{roth2022towards} in computer vision to the text domain.

\textbf{Embedding Extraction with Max Pooling.}
Given an input text, we first obtain contextualized representations using pre-trained language models such as BERT~\cite{devlin2019bert}. Since tokenizers often split words into multiple subword units, we need to aggregate subword embeddings to obtain word-level representations for alignment with token-level annotations.

While existing approaches always use only the first subword's embedding or average pooling across subwords, these methods may lose critical anomaly signals distributed across different subword positions. To maximize the anomaly signal for each token, we propose using \textbf{max pooling} to aggregate subword embeddings. Specifically, for a word split into $n$ subwords with embeddings $\{\mathbf{e}_1, \mathbf{e}_2, \ldots, \mathbf{e}_n\}$, the token-level embedding is computed as:
\begin{equation}
\mathbf{e}_{\text{token}} = \max(\mathbf{e}_1, \mathbf{e}_2, \ldots, \mathbf{e}_n)
\end{equation}
where the max operation is applied element-wise across each embedding dimension. This aggregation strategy ensures that the strongest signal from any subword position is preserved, which is particularly important for anomaly detection where abnormal patterns may manifest in specific subword positions.

\textbf{TokenCore: Token-Level Anomaly Detection.}
The core idea of our method is to construct a memory bank of normal token embeddings and measure anomalies based on distance to the nearest normal neighbor. During training, we extract word-level embeddings from all normal samples in the training set and store them in a memory bank $\mathcal{M} = \{\mathbf{m}_1, \mathbf{m}_2, \ldots, \mathbf{m}_N\}$, where $N$ is the total number of tokens in the normal training data.

\textbf{Token-Level Scoring.} At test time, for each token with embedding $\mathbf{e}_{\text{test}}$, we compute its anomaly score as the distance to its nearest neighbor in the memory bank:
\begin{equation}
s_{\text{token}} = \min_{\mathbf{m} \in \mathcal{M}} d(\mathbf{e}_{\text{test}}, \mathbf{m})
\end{equation}
where $d(\cdot, \cdot)$ denotes the distance metric (e.g., Euclidean distance). A larger distance indicates higher anomaly score, as the token is farther from any normal token pattern in the embedding space.

\textbf{Document-Level Aggregation.} To obtain document-level anomaly scores, we aggregate token-level scores across all tokens in the document. We use mean aggregation:
\begin{equation}
s_{\text{doc}} = \frac{1}{T}\sum_{i=1}^{T} s_{\text{token}_i}
\end{equation}
where $T$ is the number of tokens in the document. This aggregation provides a holistic measure of document anomaly while maintaining interpretability through token-level scores.

\section{Experiment}

\vspace{-10pt}
\begin{table}[!htbp]
\caption{Evaluation results across 7 anomaly detection methods in terms of AUROC and AUPRC. The best and second best performance are in bold and underline. }
\label{tab:results}
\resizebox{0.48\textwidth}{!}{%
\begin{tabular}{clllllll|ll}
\hline
\multicolumn{1}{l}{}   &             & \multicolumn{2}{c}{\textbf{SMS\_Spam}}                & \multicolumn{2}{c}{\textbf{Review}}                   & \multicolumn{2}{c}{\textbf{Grammar}}                  & \multicolumn{2}{|c}{\textbf{Average}}                  \\ 
\multicolumn{1}{l}{}   &             & \multicolumn{1}{c}{ROC} & \multicolumn{1}{c}{PRC} & \multicolumn{1}{c}{ROC} & \multicolumn{1}{c}{PRC} & \multicolumn{1}{c}{ROC} & \multicolumn{1}{c}{PRC} & \multicolumn{1}{|c}{ROC} & \multicolumn{1}{c}{PRC} \\ \hline
\multirow{7}{*}{\rotatebox{90}{Token}} & LOF~\cite{breunig2000lof}         & 0.5315                    & 0.0098                    & 0.8005                    & 0.0569                    & 0.5641                    & 0.0339                    & 0.6320                    & 0.0335                    \\
                       & iForest~\cite{liu2008isolation}     & {\ul 0.8157}              & \textbf{0.0341}           & 0.7052                    & {\ul 0.0768}              & 0.5024                    & 0.0303                    & 0.6744                    & {\ul 0.0471}                    \\
                       & ECOD~\cite{li2022ecod}        & \textbf{0.8186}           & {\ul 0.0289}              & 0.7035                    & \textbf{0.0872}           & 0.4457                    & 0.0268                    & 0.6559                    & \textbf{0.0476}                    \\
                       & D.SVDD~\cite{ruff2018deep}    & 0.7366                    & 0.025                     & 0.6554                    & 0.0491                    & 0.4798                    & 0.0312                    & 0.6239                    & 0.0351                    \\
                       & AE & 0.3851                    & 0.0074                    & 0.6665                    & 0.0210                    & {\ul 0.6190}               & {\ul 0.0386}              & 0.5569                    & 0.0223                    \\
                       & LUNAR~\cite{goodge2022lunar}       & 0.6734                    & 0.0139                    & {\ul 0.8254}                    & 0.0516                    & 0.6053                    & 0.0376                    & {\ul 0.7014}                    & 0.0344                    \\
                       & TokenCore    & 0.6792                    & 0.0141                    & \textbf{0.8271}           & 0.053                     & \textbf{0.6371}           & \textbf{0.0407}           & \textbf{0.7145}           & {0.0359}           \\ \hline
\multirow{7}{*}{\rotatebox{90}{Document}}   & LOF~\cite{breunig2000lof}         & \textbf{0.5948}           & \textbf{0.2276}           & 0.9082                    & 0.4107                    & 0.5236                    & 0.1933                    & 0.6755                    & 0.2772                    \\
                       & iForest~\cite{liu2008isolation}     & 0.4598                    & 0.1551                    & 0.8951                    & 0.3987                    & 0.5994                    & 0.2342                    & 0.6514                    & 0.2627                    \\
                       & ECOD~\cite{li2022ecod}        & 0.4603                    & 0.1552                    & 0.8887                    & 0.3791                    & 0.5739                    & 0.2178                    & 0.6410                    & 0.2507                    \\
                       & D.SVDD~\cite{ruff2018deep}    & 0.4814                    & 0.1633                    & 0.8572                    & 0.3371                    & 0.5548                    & 0.2294                    & 0.6311                    & 0.2433                    \\
                       & AE & 0.4576                    & 0.1566                    & 0.9075                    & 0.5247                    & 0.6403                    & 0.2547                    & 0.6685                    & 0.3120                    \\
                       & LUNAR~\cite{goodge2022lunar}       & 0.5814                    & 0.2051                    & {\ul 0.9587}                    & {\ul 0.7473}                    & \textbf{0.6578}                    & \textbf{0.2628}                    & 0.7326                    & 0.4051                    \\
                       & TokenCore    & {\ul 0.5859 }                  & {\ul 0.2072}              & \textbf{0.9594}           & \textbf{0.7495}           & {\ul 0.6553}              & {\ul 0.2624}              & \textbf{0.7335}           & \textbf{0.4064}           \\ \hline
\end{tabular}%
}
\end{table}

\vspace{-10pt}
\subsection{Experimental Setup}

We compare TokenCore against 6 anomaly detection methods and use AUROC and AUPRC as evaluation metrics. All the methods operate on BERT-base-uncased embeddings with max pooling aggregation. We use 50\% normal documents as training data, the rest of data are set as testing including both normal and anomaly.

\subsection{Results}

Table~\ref{tab:results} presents the performance comparison at both token and document-levels, the proposed TokenCore ranks first in average AUROC at both level. At the token-level, ECOD and iForest achieve the top two positions on SMS\_Spam, while TokenCore ranks first on both Review and Grammar datasets. Deep learning methods (DeepSVDD, AutoEncoder) consistently rank in the bottom half across all three datasets. The performance variance across datasets is substantial, with Review dataset showing AUROC values above 0.80 for top-performing methods, while Grammar remains below 0.65 for all methods. At the document-level, performance metrics show notable improvements compared to token-level, particularly for AUPRC. On the Review dataset, both LUNAR and TokenCore exceed 0.95 AUROC, while maintaining AUPRC above 0.74. The relative ranking of methods shifts between token and document-levels, for instance, LOF ranks first on SMS\_Spam at document-level but performs poorly at token-level.
%
%
The performance patterns across datasets reflect the characteristics of pre-trained language models like BERT, which are trained on semantic similarity objectives. On the Review dataset, semantic anomalies (negative sentiment) create clear separations in the embedding space, resulting in consistently high detection performance across methods. However, for character-level anomalies in SMS\_Spam and grammatical errors in Grammar, the embedding representations tend to smooth out these surface-level variations, as the models prioritize semantic content over form. This explains why TokenCore does not achieve the best performance on SMS\_Spam at token-level, where statistical methods like ECOD and iForest that operate on distribution deviations are more suitable for detecting orthographic anomalies. Character corruption represents surface-form violations rather than semantic deviations, suggesting that the choice of detection method should be guided by anomaly type: statistical approaches for character-level anomalies, and memory-bank methods like TokenCore for semantic and contextual anomalies. This limitation is less pronounced at document-level where aggregation provides more robust signals.

\vspace{-10pt}

\section{Challenges}

\textbf{Embedding Methods' Limitations.} Pre-trained language models are optimized for semantic similarity rather than anomaly detection. As observed in our experiments, these models effectively capture semantic deviations but tend to smooth out orthographic and syntactic anomalies. Developing embedding methods specifically designed for text anomaly detection, or adapting existing models to be sensitive to diverse anomaly types, remains an open problem.

\textbf{Multi-Granularity Aggregation.} The aggregation from token-level to document-level scores presents a non-trivial challenge. Simple strategies like mean pooling may dilute strong anomaly signals from individual tokens, while max pooling may over-emphasize isolated false positives. Designing principled aggregation methods that preserve anomaly signals across granularities while maintaining robustness requires further investigation.

\textbf{Subword-to-Word Alignment.} While max pooling preserves the strongest signal among subwords, it discards information about subword interactions and ordering. For certain anomaly types, particularly morphological errors, the relationship between subwords may carry important diagnostic information. Exploring alternative aggregation strategies that capture both individual subword signals and their compositional properties remains an open challenge.

\textbf{Generalized Anomaly Detection.} Current approaches in text anomaly detection focus on narrow, well-defined tasks such as spelling correction, hate speech detection, or fake news identification. However, real-world applications require detecting diverse anomaly types simultaneously without task-specific supervision. Bridging the gap between specialized detectors and a unified framework capable of identifying arbitrary violations of community norms poses a significant challenge for future research.

\textbf{Extreme Class Imbalance at token-level.} Unlike document-level detection where anomaly ratios are typically moderate, token-level detection faces severe class imbalance. A single anomalous document may contain only a few anomalous tokens among hundreds of normal ones, resulting in extremely skewed distributions. This imbalance poses challenges for both training and evaluation, as standard metrics may be dominated by the majority class and detection methods may struggle to identify rare anomalous tokens without generating excessive false positives.

\textbf{Computational Complexity.} Token-level detection is significantly more computationally demanding than document-level approaches. With $n$ training documents averaging $k$ tokens each, the token-level memory bank contains $nk$ embeddings compared to only n embeddings at document-level. For a test document with $m$ tokens, this requires $O(nkm)$ distance computations versus $O(n)$ for document-level methods—a difference that scales quadratically with both corpus size and document length. Approximate nearest neighbor techniques (e.g., FAISS, LSH, KD-trees) can reduce search complexity to sub-linear time, though at the cost of potential accuracy loss. This speed-accuracy trade-off warrants careful consideration in practical deployments where memory banks may contain millions of token embeddings.

\vspace{-10pt}
\section{Conclusion}

In this paper, we introduce token-level text anomaly detection, a novel paradigm that enables fine-grained localization of anomalies within text. We also propose TokenCore that detects anomalies at both token and document-levels through nearest neighbor matching in the embedding space. Our experiments on three benchmark datasets spanning character corruption, semantic anomalies, and grammar errors demonstrate that Tokencore gets better performance than other 6 baselines. This work opens new directions for interpretable and precise text anomaly detection, moving beyond document-level anomaly detection toward understanding what specific elements make text anomalous.
\vspace{-5pt}

\begin{acks}
This work was supported by National Natural Science Foundation of China (NSFC)(Grant No.62476047), 
the Guangdong Provincial 
Key Laboratory of Mathematical and Neural Dynamical
Systems(Grant No.2024B1212010004), the Cross
Disciplinary Research Team on Data Science and Intelligent
Medicine(Grant No. 2023KCXTD054).
\end{acks}
\vspace{-5pt}
\bibliographystyle{ACM-Reference-Format}
\bibliography{main}

\end{document}